# Automatic Number Plate Recognition using Random Forest Classifier

Zuhaib Akhtar[1] and Rashid Ali[2]

[1] Undergraduate Student, Department of Computer Engineering
Aligarh Muslim University, Aligarh, India
`akhtarzuhaib@gmail.com`

[2] Associate Professor, Department of Computer Engineering
Aligarh Muslim University, Aligarh, India
`rashidaliamu@rediffmail.com`

**Abstract.** Automatic Number Plate Recognition System is a mass surveillance embedded system that recognizes the number plate of the vehicle. This system is generally used for traffic management applications. It should be very efficient in detecting the number plate in noisy as well as in low illumination and also within required time frame. This paper proposes a number plate recognition method by processing vehicle's rear or front image. After image is captured, processing is divided into four steps which are Pre-Processing, Number plate localization, Character segmentation and Character recognition. Pre-Processing enhances the image for further processing, number plate localization extracts the number plate region from the image, character segmentation separates the individual characters from the extracted number plate and character recognition identifies the optical characters by using random forest classification algorithm. Experimental results reveal that the accuracy of this method is 90.9 %.

**Keywords:** Automatic Number Plate Recognition System, Edge Detection, Character Recognition, Random Forest Classifier, Ensemble Learning

## 1 Introduction

As the number of vehicles has increased considerably during the recent years, more and more attention is required on advanced, efficient and accurate intelligent transportation system (ITSs). One of the important technique used in ITS is Automatic Number Plate Recognition (ANPR) System. It was invented way back in 1979 at the Police Scientific Development Branch in the UK. However, ANPR did not become widely used until new developments in software and hardware during the 1990s [1]. ANPR is a computer vision technology that recognizes the vehicle's number plate without direct human intervention. This system captures the image of the vehicle and extracts the characters of the number plate. These extracted characters then can be searched in the database to identify the owner of the vehicle.

Hence it can be used in traffic management applications like automatic gate control for authorized/non-authorized vehicles, entrance admissions in toll systems, monitoring of traffic violations, borders crossing control and premises where high security is needed, such as the Parliament building.

The ANPR system should work under noisy conditions and low contrast. There are various methodologies used in ANPR system. The proposed methodology consists of following steps: Pre-Processing, localization of number plate using edge detection, character segmentation and character recognition. Literature survey, proposed methodology and results are discussed in subsequent sections.

## 2 Literature Review

Number Plate recognition has garnered a lot of attention from the research community. One of the important characteristics of research in number plate recognition is that the research is restricted locally as number plate tends to be different for different regions. There is no standardization of number plate. For example, each state in US has a different number pate. License plate extraction is the most important step in the number plate recognition phase. Hontai and Koga proposed a method where no prior knowledge of position and shape is required to extract characters [2]. In this method Gaussian filters were used. In order to locate number plate, neural network based filters and post processor were used [3]. Neural networks analyzed a portion of image to decide whether that portion contains number plate or not. Gabour filters were used in this method. Becerikli et. al used colors of the number plate to extract the number plate from image. In this method neural network were used for obtaining the pixel values of number plate [4]. Saqib, Asad and Omer used Hough transform based technique to extract number plate [5]. Binarization with Sauvola method and moving window were used to detect the number plate by Chang et. al [6]. After obtaining the number plate, next step was to extract characters of the number plate. Panchal et. al, used Harris corner and connected component based method to segment the characters [7]. Haar-like features and AdaBoost algorithms were applied feature extraction was reported in another study [8]. Gabor wavelet transform and local binary operator were also used. Liu and Lin used both supervised k-means and support vector machine to classify characters [9]. In the proposed method edge detecting based on Sobel vertical edge detection is used to identify the potential number plate region. Vertical projection based technique is used to segment individual characters. Random Forest Classifier, which is based on ensemble of decision trees, is used to recognize characters of number plate.

## 3 Methodology

For the present study following methodology has been used as discussed in subsequent sections and is represented in Fig.1.

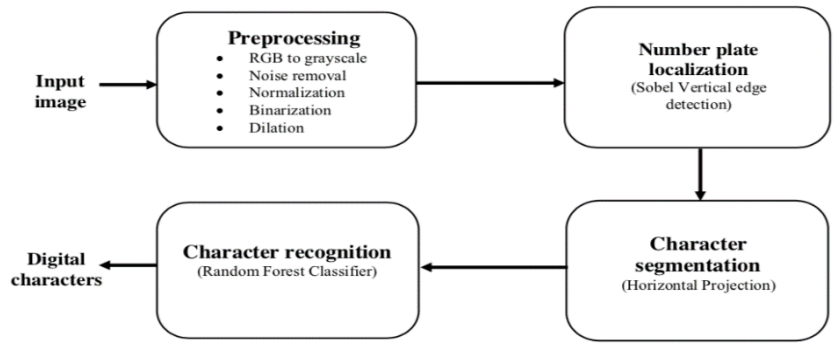

**Fig.1.** Methodology for Number Plate recognition

### 3.1    Pre-Processing

Input image suffers from many factors like noise, distortion and lack of exposure. To minimize these factors Pre-Processing is required on the input image and hence processing of image becomes easy and computationally fast. Every stage in Pre-Processing is shown in Fig.2.

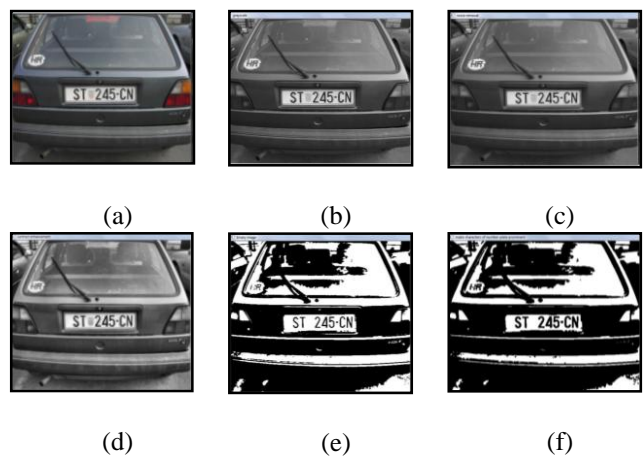

**Fig.2.** Pre-Processing steps for an image: (a) Input RGB image (b) Gray scale image (c) Noise removal (d) Contrast enhancement (e) Binary image (f) Dialated image

#### 3.1.1    Converting RGB (Red Green Blue) Image to Grayscale Image

RBG image is converted to grayscale image for two reasons as shown in Fig.2(b):

• RGB images are computationally intensive to process as compared to greyscale images simply because RGB images have three separate channels for red, blue and green values for a pixel whereas grayscale images have only single channel that represents the intensity of a pixel.

- Standard number plate has only two colors i.e., white and black. So there is no need to have the whole spectrum of colors in the image.

### 3.1.2  Noise Removal

Bilateral filter is used for removing noise (Fig.2c), kernel size 5, from image as it is very effective in removing noise while keeping the edges of characters sharp. Filtering is the process in which each pixel value in the image is replaced by the average weighted sum of the pixels nearby. Number of pixels taking part in weighting average is decided by the size of kernel. Kernel size of 5 is efficient for removing noise as size greater than 5 are slow and size of kernel less than 5 in ineffective in removing noise in our case.

### 3.1.3  Normalization

Increasing the contrast helps in cases where illumination is low. Increase in contrast increases the separation between colors which helps in separating the black characters from the white background on the number plate (Fig.2d).  Contrast Limited Adaptive Histogram Equalization (CLASH) is used for increasing the contrast. This technique divides the image into small blocks called tiles. Tile of size 8×8 was used and histogram equalization is applied on these tiles independently. This keeps the information in regions which is too much exposed to brightness. Some examples which is significantly helped by contrast enhancement is shown in Fig. 3, that clearly shows it makes character on the number plate prominent.

| Before contrast enhancement | After contrast enhancement | Before contrast enhancement | After Ccontrast enhancement |

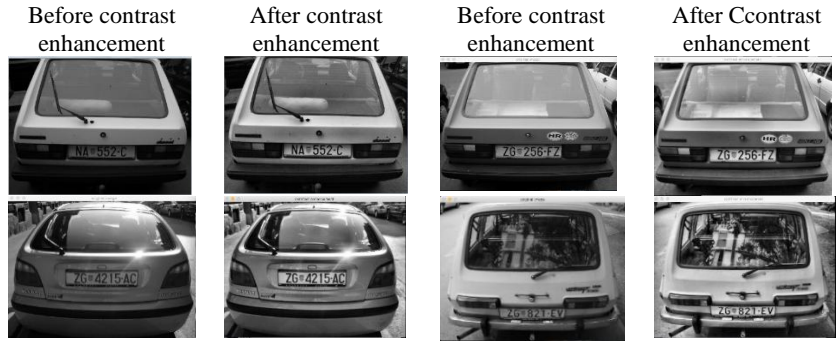

**Fig.3.** Normalization

### 3.1.4  Converting to Binary Image

To make the color domain of image same as that of number plate, image is converted into binary image (Fig.2e), which reduces the pixel values to only two values i.e., black (0) and while (1). This is done by choosing a threshold (128 was chosen on the scale of 255). If the value of the pixel is less than threshold then it is converted to black color (value = 0) and if the value is greater than threshold it is converted to white color (value = 255).

### 3.1.5 Dilation of Image

Dilation of image serves two purposes (Fig.2f):

- It makes the characters of the number plate bold which increases the area of the characters which in turn increases their edges. This is because of increase in area of an object that also increases the perimeter (edges) of the image. Increase in the edges of the character helps in localization of the number plate as the plate is localized by edge detection method.

- It removes the unnecessary edges around the number plate, hence prevents from capturing the "false" edges. This is because it reduces the perimeter of noise blot by combining noises nearby, hence reducing the edges of noise. This is shown in Fig. 4.

  Fig. 4(a) shows the number plate region without dilation and Fig. 4(b) shows its edges and vertical projection of the edges. Similarly, Fig. 4(c) shows the number plate region with dilation and Fig. 4(d) shows its edges and vertical projection of the edges. Fig. 4(b) clearly shows that peaks of noise is greater than peaks of noise in Fig. 4(d) which is present just below the characters of the number plate. Hence, dilation helps to reduce the edges of noise surrounding the number plate.

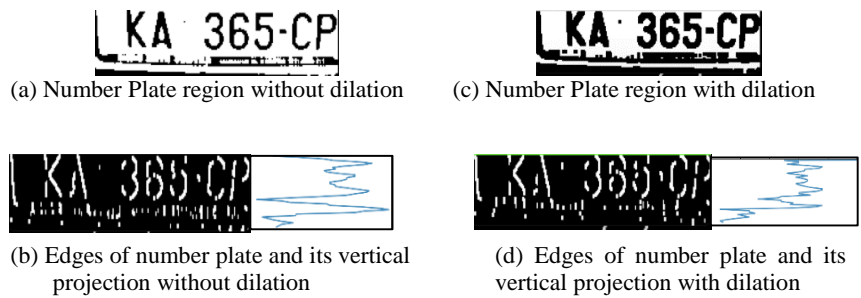

(a) Number Plate region without dilation

(c) Number Plate region with dilation

(b) Edges of number plate and its vertical projection without dilation

(d) Edges of number plate and its vertical projection with dilation

**Fig.4.** Dilation reduces the edges of noise surrounding the number plate

### 3.2 Localization of Number Plate

To localize the number plate, edge detection is used, which gives the edges of the image. Most of the edges are localized on the number plate region which are actually the characters on the number plate. This is because standard number plate has black characters on the white number plate. Therefore, edges around the characters are distinctly separated from background. Also the characters are concentrated on the number plate which forms the region of maximum localized edges. Although, the edges are present throughout the image, but maximum edges are locally concentrated on the number plate. Pre-Processing makes these edges distinct. It's because of this property, bounding rectangle is taken, which represents the number plate. It is traversed on the image which counts the number of edges it bounds. Wherever it finds the maximum localized number of edges in the bounding rectangle, that portion of image is taken as the number plate.

Canny Edge detection method was tried to localize the number plate. Although Canny edge detection is more robust edge detection method, but it detects both horizontal and vertical edges. Generally in vehicle, just above the number plate there is a back windshield of car which has horizontal structure. Similarly, just below the number plate there is bumper which again has almost horizontal structure. The number plate itself and all the structures around the number plate gives strong horizontal edges as shown in Fig. 5(a). When Canny edge detection is applied it captures these edges and hence, only some portion or incorrect number plate region is captured, which is shown in Fig. 5(c) enclosed by green colored rectangle. Fig. 5(e) clearly shows partial number plate region is captured when canny edge detection is used.

Sobel vertical edge detection removes all the horizontal edges from the image. It only captures vertical edges which is clear in Fig. 5(b). Hence, all the horizontal edges in the image are removed. The edges of characters are still recognized as they are combination of both horizontal and vertical edges. Fig. 5(d) shows the localized number plate region is detected which is accurate and is shown by enclosed green colored rectangle. Fig 5(f) shows captured area of number plate when sobel vertical edge detection is used.

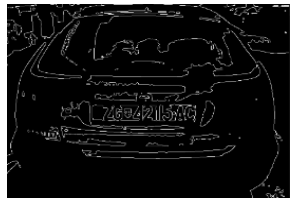

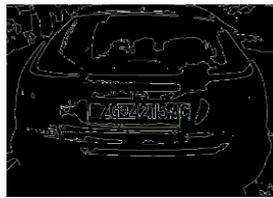

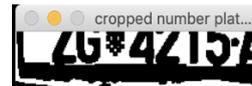

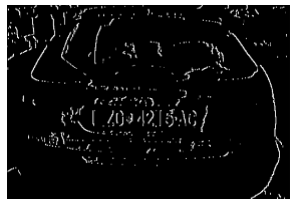

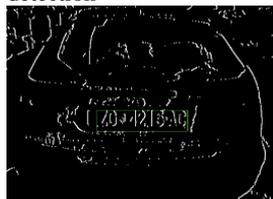

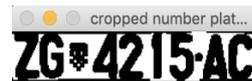

(a) Canny edge detection of vehicle

(c) Localized number plate shown by green highlighted area using Canny edge detection

(e) Localized number plate using Canny edge detection

(b) Sobel vertical edge detection of vehicle

(d) Localized number plate shown by green highlighted area using sobel vertical edge detection

(f) Localized number plate using sobel vertical edge detection

**Fig.5.** Comparision between localization of number plate using Canny edge detection and Sobel vertical edge detection

Green bounded rectangle shown in Fig.6(a) represents the maximum localized edges inside the bounding rectangle which is the number plate. That portion of image is cropped from dilated image (Fig.6b) and used for character segmentation.

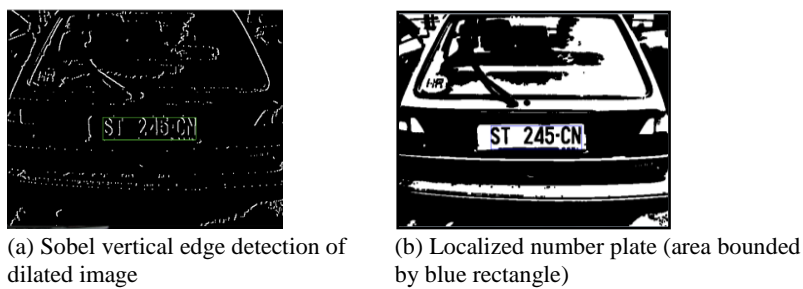

(a) Sobel vertical edge detection of dilated image

(b) Localized number plate (area bounded by blue rectangle)

**Fig.6.** Localization of number plate

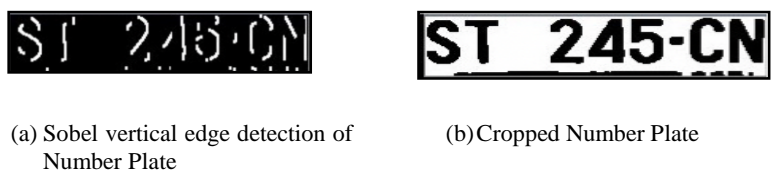

(a) Sobel vertical edge detection of Number Plate

(b) Cropped Number Plate

**Fig.7**. Localized Number plate

### 3.3    Character Segmentation

Before this stage, number plate has been localized. In this stage, localized number plate is taken as an input and the characters present in the number plate are extracted from it.

### 3.3.1    Removal of Redundant Portion of the Number Plate

Before segmentation of characters is carried out, noise surrounding the characters is removed. If there is no noise around the characters then empty white spaces of the number plate should be removed to make computation faster. These portions are removed by first applying Sobel vertical edge detection on Fig. 7(b) which gives Fig 7(a). We are only interested in vertical edges because noises below and above the characters generally have greater horizontal component than vertical component. Hence, taking vertical edges remove all horizontal components of the noise. Characters are not affected by this because they are blend of horizontal and vertical components.

These edges (white pixels as shown in Fig. 7(a) are projected on the vertical axis. Projection on the vertical axis is done by counting the number of white pixels (which represent edges) of each row and storing the value of count in an array. This array is converted into a histogram. The characters are generally represented by the highest and widest peaks in the histogram as shown in the Fig 8. If there is no noise then there is only single band of peaks representing characters.

First the numbers of bands are searched in the image. Then these bands are compared with each other on the basis of width and height of the bands.

As there is space between characters and noise, this space will separate the band of peaks of characters and noise into two separate bands. So, the band of peaks representing characters are cropped out, as they are highest and widest, hence, number plate free from noise is acquired as shown in Fig 9. After obtaining this image, it can now be sent to the stage of segmenting the individual characters.

There were some cases when the peaks of noise were higher than those of characters as shown in Fig 10. When Sobel vertical edge detection is applied on Fig. 10, peaks of noise becomes greater than that of characters as shown in Fig. 11. For such cases, these peaks of noise were eliminated due to the fact that characters represent wider band of peaks than noise and hence, number plate free from noise is obtained as shown in Fig. 12.

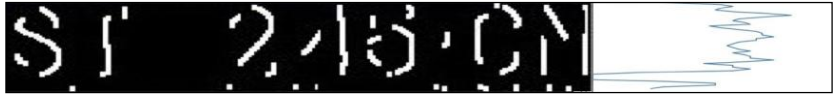

**Fig.8.** Vertical projection of edges of number plate. The histogram on the right of image shows the number of edges (white pixels) on each row. The histogram dips to zero where there are no edges (white pixels).

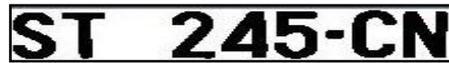

**Fig.9.** Number plate after taking the highest and widest peaks of characters which makes it free from noise (compared with Fig. 7(b) )

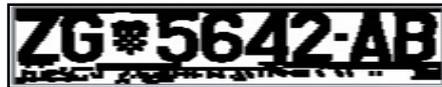

**Fig.10.** localized number plate with noise.

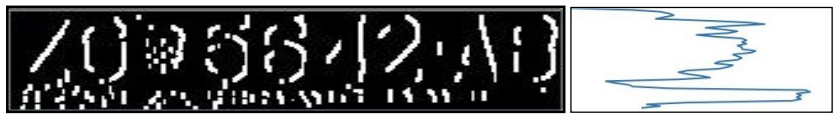

**Fig.11.** Vertical projection of edges of number plate in Fig. 10. The histogram on the right of image shows the number of edges (white pixels) on each row. Peaks of noise are greater than characters but are lesser in width.

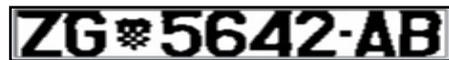

**Fig.12.** Number Plate free from noise

### 3.3.2 Segmentation of Characters

The image shown in Fig.9 is free from noise. This image is projected on the horizontal axis. This is done by summing black pixels of each column and storing it in an array. This array is plotted in a histogram as shown in Fig 13. As there are gaps between each character so there are no black pixels in between and hence the value of histogram is zero in these gaps. Non-zero values in histogram represents characters. So to get the characters out of the number plate, these individual peaks are cropped out, from the position where the peak starts to the location where the peak ends, which represents a character.

This technique is applied throughout the image which gives all the characters. These individual characters are send to the next stage for recognition (Fig 14).

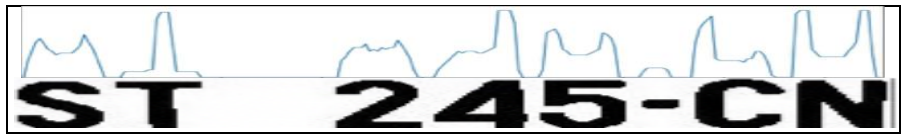

**Fig.13.** After image becomes free from noise, it is projected on the horizontal axis. Peaks of histogram represents the number of black pixels in each column of number plate. Image between these peaks are cropped which gives the individual characters.

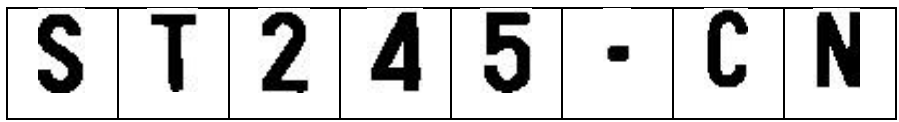

**Fig.14.** Separated characters from Number plate

### 3.4 Character Recognition

For classifying the segmented characters into the respective digits and alphabets, various machine learning classification techniques were tried, such as neural networks, k-nearest neighbor (k-NN), support vector machines (SVM) and Random Forest Classifier (RF), as No Free Lunch theorem states that there will always be data sets where one classifier is better than another.

### 3.4.1 k-NN Classification Algorithm

k-NN is a classification algorithm for supervised learning and it is non- parametric approach for classification. The value of 'k' was chosen as three, where k is number of nearest neighbor. The value of 'k' plays an important role in performance of k-NN. It is the tuning parameter of the model. k-NN groups the training images into classes, similar images are grouped together in the form of a class. When test image is provided, image is plotted on the same graph and 'k' closest training images are chosen from test image.

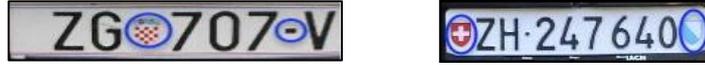

**Fig.15**. Number plates with encircled special characters and symbols

A major challenge was to classify special characters and symbols as shown in Fig 15 (encircled on number plate). These symbols appeared frequently in number plates throughout the dataset and needed to be eliminated from the output. To tackle this problem, these characters were classified into two different classes as shown in Fig. 16.

| Class - A | | Class - B | |
|---|---|---|---|

**Fig.16.** Classification of special characters

To disqualify images that belong to class A, number of black pixels were calculated. In actual characters there was healthy number of black pixels. In other words, lager area of the image was covered by the character itself. However, that is not the case with the images of Class A. Therefore a threshold (Tc) was set. Whenever total number of black pixels in an image was less than the set threshold, that image was discarded. This threshold was also used in subsequent classifiers.

For class B, this technique could not be used as they have large number of black pixels. Implementation of kNN in OpenCV returns the shortest distance between test characters and the characters that were used for training (neighbors). The more the neighbors are closer to test image, the closer the match. If this distance is very large, i.e., if the test image does not even remotely resemble any of the training images, then the image was unlikely to be a legitimate character. Again, a threshold value was set (Ts). If the distance between test image and nearest neighbors was greater than threshold, then that image was discarded.

### 3.4.2    Multi-Layer Perceptron

Multi-Layer Perceptron or Neural Network (NN) is composed of artificial neurons which mimics human brain. Neural Network consists of layers and each layer has number of neurons. Layers are categorized into three types. Input layer, hidden layers and output layers. Input layer is equals to number of features or in this case, it is individual pixles. So, it is equals to 400 (size of image is 20×20).

Number of neurons in output layer is equals to number of classes. We have 36 classes to represent digits and alphabet (10 digits and 26 alphabets). Hence, output layer has 36 neurons. In between input and output layer, neural network has hidden layers. For this problem, two hidden layers were chosen, first hidden layer has 270 neurons and second hidden layer has 150 neurons. Each neuron is generally connected by weights to all the previous layer neurons. Output of previous layer is multiplied and added to each other before it is fed to the neuron of next layer as an input.

**Table 1.** Details of important neural network parameters

| Type of Classifier | Multi-Layer-Perceptron |
| --- | --- |
| Total neurons in input layer | 400 |
| Total neurons in first hidden layer | 270 |
| Total neurons in second hidden layer | 150 |
| Total neurons in output layer | 36 |
| Learning rate | 0.0001 |
| Activation function | Logistic function |
| Maximum iterations for training set | 500 |

At that neuron, output is calculated by activation function by using input. It maps the output of neuron between 0 and 1 or -1 and 1. There are several parameters that need to be set for neural network. One is activation function which is present in each neuron. Logistic function was chosen as an activation function which maps the values between 0 and 1. Learning rate was set to 0.001. Maximum iterations was set to 500 and training stopped at 278th iteration. To eliminate characters of class-A, count threshold 'Tc' was and to eliminate characters of class-B, threshold (Ns) was set at the output of neural network to ignore those special characters. All the important parameters are listed Table 1.

### 3.4.3 Support Vector Machine

SVM is a discriminative classifier. It categorizes the data by forming the hyperplane. It works best for binary class classification. Although, SVMs can be modified for multi-class problems. It usually consists of constructing binary classifiers which distinguish between one label and rest, called one-versus-all approach or between every pair of classes, called one-versus-one approach.

**Table 2.** Details of important support vector machine parameters

| Type of Classifier | Support Vector Machine |
| --- | --- |
| Approach | One-vs-all |
| optimum cost parameter (C) | 1 |
| kernel width parameter (γ) | 'auto_deprecated' |
| degree of polynomial kernel | 3 |

One-versus-all approach was chosen for SVM. There are three important parameters that needs to be set when SVM is applied which are: kernel, kernel width parameter (γ) and optimum cost parameter (C). Polynomial kernel of degree 3 was chosen for this case. Parameter 'C' decides the size of misclassification allowed, i.e., how much one wants to avoid misclassification of each training data. Large values of 'C' will choose smaller margin hyperplane. Conversely smaller values of 'C' will force classifier to look for larger margin, even if that hyperplane misclassifies the points. This is shown in Fig 17. Generally high value of parameter 'C' is desirable but it might lead to over fitting. Parameter 'γ' affects the shape of the class dividing hyperplane. When 'γ' is high, only points near the hyperplane is taken into consideration whereas, low value of 'γ' will take far away points from hyperplane into consideration. Generally, low value of 'γ' is desirable as larger value might lead to over fitting. High value of optimum cost parameter (C=1) was chosen with low value of kernel width parameter (a default value γ = 'auto_deprecated'). To eliminate characters of class-A, count threshold 'Tc' was used and to eliminate characters of class-B, probability estimate was used by setting a threshold (Ps). All the important parameters are listed in Table 2.

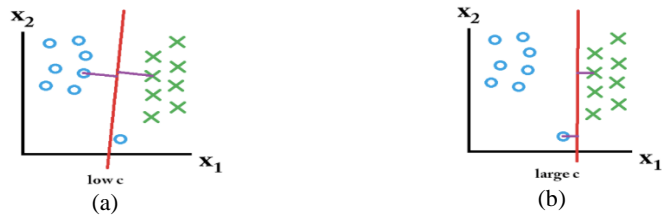

**Fig.17.** Hyperplane with low 'C' (a) and hyperplane with large 'C' (b)

### 3.4.4 Random Forest Classifier

Random forest classification algorithm that is based on ensemble of decision trees where each of the tree is based on randomly selected subset of training set. Tree consists of nodes where the decision is taken on some parameter. This forest is time trained with a method which is based on bagging.

**Table 3.** Details of important random forest parameters

| Type of Classifier | Random Forest |
|---|---|
| Number of trees (Nt) | 100 |
| Number of features in each split (Fs) | $\sqrt{number\_of\_features}$ |

Random Forest uses slightly different kind of bagging approach where a subset of features is selected for the split at node, whereas, in bagging all features are used for node split. As the result of random forest is aggregation of trees, which reduces the affect of noise present in a single tree. Hence, bagging generally increases the overall result. Random forest are inherently multiclass which can be used in our problem case. There are two important parameters that needs to be set, one is number of features in each split (Fs) and number of decision trees (Nt) in the forest. The large value of parameter 'Nt' may be unnecessary but it does not harm the model. It will definitely make the predictions stronger but might make the model slower. Parameter 'Fs' is the number of features to consider while splitting a node. It is always subset of number of features. Value of 'Nt' was chosen as 100 and 'Fs' was set as square root of number of features for the model. To eliminate characters of class-A, count threshold 'Tc' was used and to eliminate characters of class-B, probability estimate was used by setting a threshold (Pe). All the important parameters are listed in Table 3.

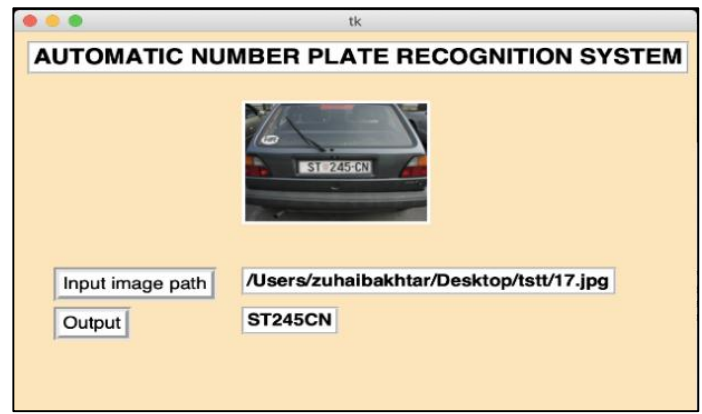

**Fig.18.** Final ouput of Number Plate Recognition System

## 4.    Results and Discussions

Automatic Number Plate Recognition System was written in Python (3.6.5). Libraries such as OpenCV (3.4.2), numpy (1.14.0), scikit-learn (0.19.1) and matplotlib (2.2.3) were used. Graphical User Interface was created using tkinter (8.6.8). Out of 350 images in the dataset (images of Croatian vehicles), 100 were used for testing, 220 for training and rest was used for validation to fine tune hyper parameters of the various models used. All the segmented characters used for training were resized to 20×20. Different models were trained by training dataset. Characters were first extracted from the number plate by character segmentation and then these characters were used for training the various classification models. Final output of the Number Plate Recognition System is shown in Fig. 18 which shows recognized characters. Accuracy of various models is shown in Table 4.

### 4.1    Testing

The final character accuracy for k-NN was found to be 83.40%. This classifier was found to be more prone to making mistakes between visually ambiguous characters, such as '8' and 'B' ,'I' and '1' ,'O' and 'D' ,'G' and '6'. The results were near perfect for characters that are not optically ambiguous. Time taken by k-NN classifier was 0.3 seconds to give the output of an image.

**Table 4.** Accuracy of various models tested

| Classifier | Accuracy |
|---|---|
| k-NN | 83.4% |
| Neural Network | 89.47% |
| SVM | 87.5% |
| Random Forest | 90.9% |

In case of Neural Network, character accuracy was found to be 89.47%. This classifier was found to be prone to making mistakes between 'O' and 'G' and very rarely between '2' and 'Z'. Time taken by Neural Network was 0.23 seconds to give the output of an image.

In case of SVM classifier, character accuracy was found to be 87.50%. This classifier sometimes was found to be making mistakes between '8' and 'B' , 'I' and '1' , 'G' and '6', although it gave correct output for 'O' and 'D' most of the time. Time taken by SVM was 0.31 seconds to give the output of an image.

In case of Random forest classifier, character accuracy was found to be 90.9%. The classifier completely removed ambiguity between 'G' and '6' and reduced the error in detecting characters '8' and 'B', 'I' and '1' to a greater extent than SVM. Time taken by Random Forest was 0.35 seconds to give the output of an image.

### 4.2    Discussion

There are various reasons why RF works better than SVM in this problem. First, Random forest is naturally a multi class classifier whereas SVM is binary classifier. For SVM to work in multi class problem, it is reduced to multiple binary class problem. Still, results show random forest outperforms SVM because it is intrinsically a multi class classifier. Second, images are inherently noisy even if Pre-Processing is performed on it. Noise resistant classier should perform better in this case. This claim is backed up by the results where RF performs better than SVM. RF gives the overall result by taking the consensus result of different trees present in the classifier. Hence even if some trees get trained on noisy input, overall result is expected to give the desired output. Also, RF actually don't take long to train, especially if you do so in parallel, something one cannot do with SVM. Neural networks are also noise tolerant; hence it gives decent accuracy. Still, it does not beat the accuracy of random forest which brings the robustness of ensemble of decision trees which is based on kind of bagging approach.

## 5.    Conclusion

The algorithm for number plate recognition has been proposed. Pre-Processing, Number plate localization, Character segmentation and Character recognition are the steps used in the algorithm. Pre-Processing includes converting RGB image to grayscale image, removing noise by using Bilateral Filter, then increasing the contrast of image using CLASH, converting the image to a binary image and finally dilating the image.  Number plate localization extracts the number plate region from the image using Sobel vertical edge detection. It is also shown in this proposed work that Canny edge detection does not work as effectively as Sobel vertical edge detection because of its intrinsic characteristics. Character segmentation first removes the redundant portion of number plate which might hinder the extraction of characters and then segments the individual characters from the extracted number plate. Character recognition recognizes the individual optical characters by using Random Forest Classification Algorithm. For character recognition, Random Forest works best among several classification methods because it is based on ensemble of classifiers; precisely decision trees.